\documentclass[11pt]{article}
                  \usepackage{llncsdoc}
\usepackage[utf8]{inputenc}
\usepackage[T1]{fontenc}
\usepackage{times}
\usepackage{url}
\usepackage{latexsym}
\usepackage{wrapfig}
\usepackage{graphicx}
\usepackage{multirow}
\usepackage[margin=25mm,bottom=25mm,left=25mm,right=25mm]{geometry}
\usepackage[romanian, english]{babel}
\usepackage{combelow}
\usepackage{newunicodechar}
\newunicodechar{Ș}{\cb{S}}
\newunicodechar{ș}{\cb{s}}
\newunicodechar{Ț}{\cb{T}}
\newunicodechar{ț}{\cb{t}}
\author{ Liviu P. Dinu, Ana Sabina Uban
\\ Faculty of Mathematics and Computer Science,  \\ Human Language Technologies Research Center, \\ University of Bucharest
\\{\tt liviu.p.dinu@gmail.com}, {\tt ana.uban@gmail.com}}
\date{\vspace{-5ex}}
\title{Analyzing Stylistic Variation across Different Political Regimes}
\begin{document}

\maketitle
\begin{abstract}

In this article we propose a stylistic analysis of texts written across two different periods, which differ not only temporally, but politically and culturally: 
communism and democracy in Romania.
We aim to analyze the stylistic variation between texts written during these two periods, and determine at what levels
the variation is more apparent (if any): at the stylistic level, at the topic level etc.
We take a look at the stylistic profile of these texts comparatively, by performing clustering and classification experiments on the texts,
using traditional authorship attribution methods and features. To confirm the stylistic variation is indeed an effect of the change in
political and cultural environment, and not merely reflective of a natural change in the author's style with time, we look at various stylistic metrics
over time and show that the change in style between the two periods is statistically significant.
We also perform an analysis of the variation in topic between the two epochs, to compare with the varation at the style level.
These analyses show that texts from the two periods can indeed be distinguished, both from the point of view of style and from that of semantic content (topic).

\end{abstract}
\section{Previous Work and Motivation}
\label{sec-1}

Automated authorship attribution has a long history (starting from the early 20th century \cite{mendenhall1901mechanical}) and has since then been
extensively studied and elaborated upon. The problem of authorship identification is based on the assumption 
that there are stylistic features that can help distinguish the real author of a text from any other theoretical author.
This said set of stylistic features was recently defined as a linguistic fingerprint
(or stylome), which can be measured, is largely unconscious, and is constant \cite{van2005new}. 
One of the oldest studies to propose an approach to this problem is on the issue of the \emph{Federalist Papers},
in which \cite{mosteller1963inference} try to determine the real author of a few of these papers which have
disputed paternity. This work remains iconic in the field, both for introducing a standard dataset and for
proposing an effective method for distinguishing between the authors' styles, that is still relevant to this day,
based on function words frequencies. Many other types of features have been proposed and successfully used 
in subsequent studies to determine the author of a text. These types of features generally contrast with the 
content words commonly used in text categorization by topic, and are said to be used unconsciously and harder
to control by the author. Such features are, for example, grammatical structures \cite{baayen1996outside},
part-of-speech n-grams \cite{koppel2003exploiting}, lexical richness \cite{tweedie1998variable}, or even the more general
feature of character n-grams \cite{kevselj2003n,dinu2008authorship}. Having applications that go beyond
finding the real authors of controversial texts, ranging from plagiarism detection to forensics to security,
stylometry has widened its scope into other related subtopics such as author verification (verifying whether a text
was written by a certain author) \cite{koppel2004authorship}, author profiling (e.g. gender or age prediction),
author diarization, or author masking (given a document, paraphrase it so that the original style does not match that of its original author anymore).

In this work we attempt to explore a slightly different issue, strongly related with the stylistic fingerprint of an 
author, namely if an author preserves his stylome across different time periods, and across political and cultural environments.
More specifically, we want to see if we can discriminate between texts written by the same author under a communist regime, as compared to under democracy.

For this analysis, we chose the works of Solomon Marcus (1925-2016), one of the most prominent scientists and men of culture of modern Romania,
and a very prolific author whose work spans a large period of time, with a consistent amount of work both before and after the fall of communism in Romania.
As a scientist, he was active in an impressive range of different fields, such as mathematics, computer science, mathematical and computational linguistics,
semiotics etc, and published over 50 books translated in more than 10 languages,
and about 400 research articles in specialized journals in almost all European countries. 
He is one of the initiators of mathematical linguistics \cite{marcus1971introduzione} and of mathematical poetics \cite{marcus1973mathematische},
and has been a member of the editorial board of tens of international scientific journals covering all
his domains of interest. As a man of culture,
he has written an equally impressive amount of texts, having as preferred topics mathematical education, culture, science, children, etc.
His wide interests and complex personality have left deep imprints in the Romanian scientific and cultural world.

We chose these works and this period in Romanian history as a particularly interesting study case for stylistic variation.
Marcus' essays \cite{marcus2012rani} were written over a period spanning almost half a century:
22 years of communist regime and 27 years of democracy (after the fall of communism in Romania in 1989).
The significant changes in Romania's cultural life with the fall of communism motivate such an analysis beyond a simple temporal
analysis of texts: in the pre-democracy period, the communist norms demanded an elaborated writing style,
whereas the list of officially approved topics was much more limited.
On the other hand, with the fall of communism, we expect that the author would
would be able to radically change his style to express his ideas freely.

Previous work on the same corpus \cite{dinu-dinu-dumitru:2017:RANLP} focused on a lexical quantitative analysis of Marcus' works,
uncovering some differences in word usage between the communist and the democracy period. We continue this work by performing
more robust analyses using various metrics and classification experiments to show that the texts written during the two periods are indeed
distinguishable with a reasonable accuracy.

\section{The Corpus}
\label{sec-2}

For the purposes of the analysis presented here, we used a collection of Solomon Marcus' non-scientific texts.
The whole collection of Marcus’ non-scientific publications is available in print,
in six volumes entitled "Răni deschise" (\emph{Open wounds}), of which we analyze the first four: two containing texts written
before the fall of communism, and the other two with texts written under democracy.
The first volume comprises 1256 pages of texts, conferences and interviews, from 2002 to 2011 (and a few published before 1990).
The second volume "Cultură sub dictatură" (\emph{Culture under dictatorship}) of 1088 pages and the third "Depun Mărturie" (\emph{I testify}), of 668 pages,
is a collection of texts published between 1967 and 1989. The fourth volume "Dezmeticindu-ne" (\emph{Awaking}), of 1030 pages, contains texts from
 the period immediately following the Romanian Revolution (December 1989) to the year 2011.

Before the texts could be used for automatic analysis, some pre-processing was involved, including extracting the texts from PDF files into
text form, and splitting the corpus into individual essays.
We also parsed the table of contents in order to label each text with its year of publication. Additionally, we excluded some texts
that would interfere with the experiments, such as texts in languages different than Romanian (most of the texts are in Romanian),
and interviews (which also contain lines of the interviewer, that should not be considered when analysing text authored by Solomon Marcus).

\section{Methodology}
\label{sec-3}

In order to analyze the style of Marcus' work and its variation, we perform a series of classification and clustering experiments, as well as a temporal
analysis of the evolution of the author's style. We look at various features to capture both style and semantic content, with a focus on metrics that 
that characterize the stylistic fingerprint of the author.

We perform various classification and clustering experiments, as well as visualisations complementing them, for which we experiment first with simple bag of words models.
To capture the stylistic fingerprint of the author, we resort to one of the most commonly accepted indicators of personal style of an author,
namely the distribution of functional words, such as pronouns, determiners, prepositions, etc, which were successfully used in many previous studies to solve authorship attribution problems.
For comparison, we perform a parallel experiment where we use as features
only the content words, assuming they are descriptive of the semantic content of the texts, as opposed to their style. 

Aside from these feature sets, we also use more sophisticated features in some additional classification experiments: various stylistic metrics (such as readability or
lexical richness) to capture the stylistic variation, as well as features resulted from topic modelling to capture the semantic content. These features will be described
in more detail in the chapters dedicated specifically to the classification experiments.

\subsection{Clustering Experiments}
\label{sec-3-1}

Processing the text involves extracting all function words from each essay, and representing each text as a vector of frequencies of function words.
We use a curated list of 120 functional Romanian words \cite{dinu2012pastiche}. 
On these vectors we then apply similarity metrics in order to cluster texts based on their stylistic profile.
The distance we have chosen is rank distance \cite{dinu2003classification}, \cite{popescu2008rank},
an ordinal distance which was successfully previously used in other problems of authorship \cite{dinu2008authorship}, \cite{dinu2012pastiche}.

For our purposes, we applied a hierarchical clustering algorithm, and plotted the resulted dendrogram, as shown in Figure 1.
Texts from each volume are marked with the prefix \emph{RD<volume\_number>}.
The intention is to see whether texts from the same period are clustered together: in this case, we would expect to see texts
from volumes 1 and 4 (communism) appear close together, and texts from volumes 2 and 3 (post-communism) appear in a separate group.

An additional processing step we performed before applying the clustering algorithm was to group the essays into longer texts, while
keeping them separated into volumes (texts from different volumes/periods were not grouped together). For each volume,
batches of 10 essays were grouped toghether in 1 text.
This results in a fewer number of texts, which can be more easily plotted on a dendrogram, and additionally should contain less
noise than the very short original texts.

\begin{figure}[htb]
\centering
\includegraphics[width=1.0\linewidth]{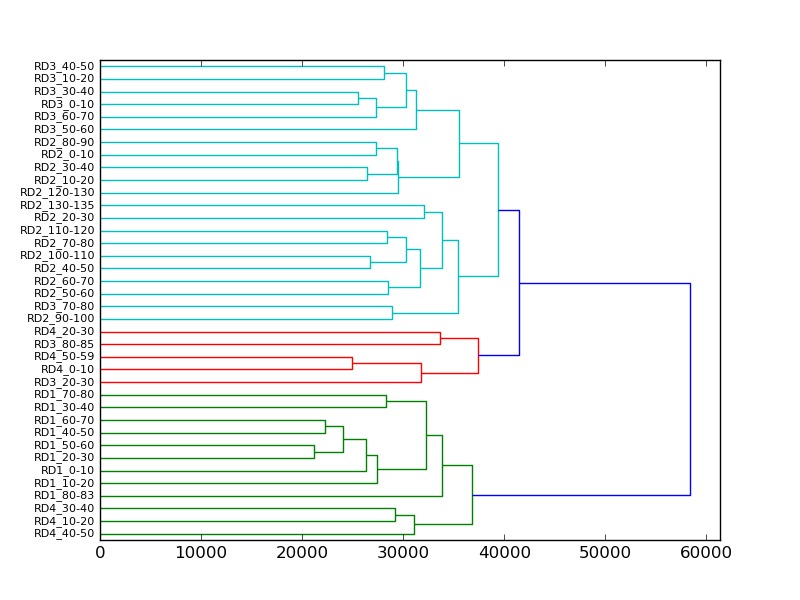}
\caption{Style dendrogram: Hierarchical clustering of function word vectors}
\end{figure}

\subsection{Dimensionality Reduction and Visualization}
\label{sec-3-2}

In addition to the clustering experiments, we perform an experiment where we plot the essays in the 4 volumes in 2-D space,
by initially applying dimensionality reduction (principal component analysis) on the texts, represented as function word frequency
vectors, as described above. The resulted plot is shown in Figure 2.

\begin{figure}[htb]
\centering
\includegraphics[width=1.0\linewidth]{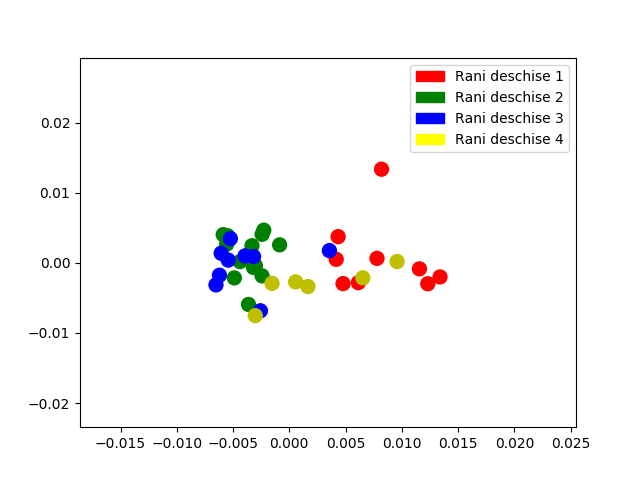}
\caption{Style space: 2-D visualization of function word vectors}
\end{figure}

In a separate experiment, we take a brief look at the content profile of the texts (as opposed to the stylistic one), by performing the same
analysis on content words instead of function words. We extract all content words from all the (grouped) texts, and represent the texts as 
vectors of content words frequencies. We then apply PCA on the resulted vector space, and plot the 2-dimensional representations of
the texts, as shown in Figure 3. Although this is still a very crude analysis, the result should provide an insight into the content similarities and dissimilarities
between the texts in the 4 volumes.

We also perform a version of the clustering experiments on the same vectors of content words, with the same methodology that was applied in the
stylistic analysis: the rank distance similarity is applied on the word vectors, and the distances are used in a hierarchical clustering algorithm,
which results in the "content" dendrogram in Figure 4.

\begin{figure}[htb]
\centering
\includegraphics[width=1.0\linewidth]{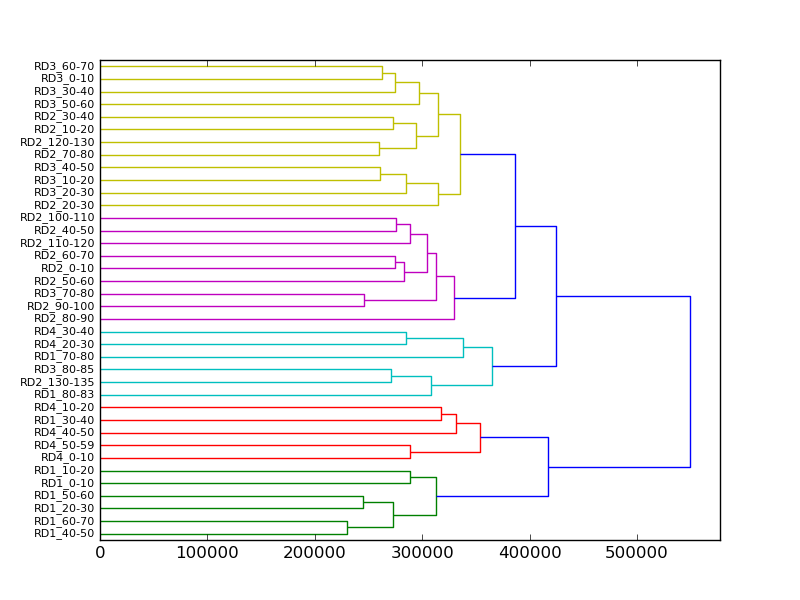}
\caption{Topic dendrogram: Hierarchical clustering on content word vectors}
\end{figure}

\begin{figure}[htb]
\centering
\includegraphics[width=1.0\linewidth]{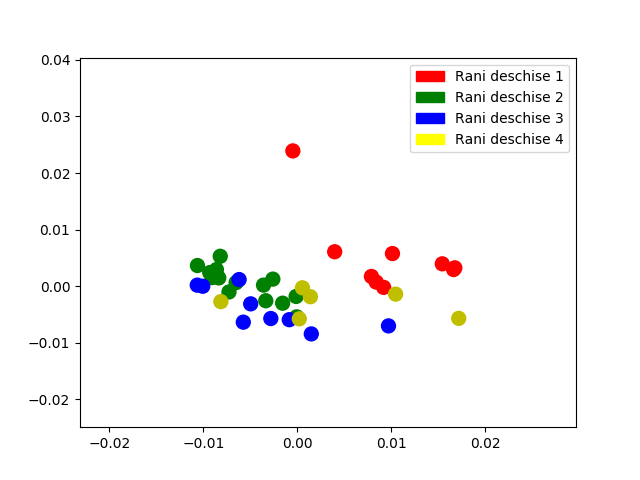}
\caption{Topic space: 2-D visualization of content word vectors}
\end{figure}

\subsection{Classification Experiments}
\label{sec-3-3}

In order to have a more conclusive result regarding the differences between the texts written in the two epochs and the levels at which
they occur, we also performed some classification experiments where we try to predict for each essay if it was written before or
after the fall of communism, as well as to which specific volume it pertains.

For classifying the texts we used an SVM classifier with a linear kernel, and tested its performance in a series of leave-one-out experiments.

\subsubsection{Features}
\label{sec-3-3-1}

We used various sets of features, in turn specific to style and to semantic content, in order to conclusively characterize the nature of the 
evolution of Marcus' writing over time and across the two periods.

As stylistic features, we experimented with 2 setups: first a bag-of-words model using only functional words; and secondly a set of stylistic markers:

\begin{itemize}
\item \textbf{Automated Readability Index}, as defined in \cite{senter1967automated}, computed as:
\end{itemize}
$$ ARI = 4.71\frac{c}{w} + 0.5\frac{w}{s} - 21.43 $$
where \emph{c}, \emph{w} and \emph{s} represent, respectively, total number of characters, words and sentences in the text

\begin{itemize}
\item \textbf{Lexical richness}, computed as the number of unique lemmas divided by the total number of tokens in the text:
\end{itemize}

$$ LR = \frac{number\ of\ unique\ lemmas}{total\ number\ of\ tokens} $$

\begin{itemize}
\item \textbf{Average word length}
\item \textbf{Average sentence length}
\end{itemize}

To measure the variation in semantic content, we performed two classification experiments. In the first experiment we used a simple bag-of-words model, this time using only content words
as features. For the second experiment, we used topic modelling to extract topics in each article, that we then used as features in the classifier. To extract the topics,
we used a Latent Dirichlet Allocation model, configured to generate 3 latent topics for each document. 

Results are shown in Table 1 and Table 2: in Table 1 we report the results of trying to classify the texts into the two distinct periods, while Table 2 shows the
results of classifying each text into the specific volume to which it pertains.

\subsection{Temporal variation of stylistic metrics}
\label{sec-3-4}

In order to show that the stylistic variation discovered in the experiments described in the previous chapter can indeed be attributed to the change of
political regime, and not simply to the passing of time and the natural gradual change in the author's style, we also look at how the various stylistic metrics
vary over time at a finer level.

Figures 5 to 8 show the evolution of each stylistic metric described in the previous chapter from year to year. To achieve this, we computed each metric for every article
in the corpus, then for each year we averaged the values obtained for the articles written during that year. To reduce the effect of noise and variation in number
of articles for each year, we smooth the obtained values by
applying a moving average over the articles in each year with a window of 10 years.

In order to confirm the variation in style is more pronounced around the year 1990 (that marks the beginning of democracy in Romania),
we performed a series of experiments in which we attempted temporal splits of the article set into two periods using each year as a separator,
and computed the statistical significance of the difference shown at the level of each metric for each of these splits. We assume that the more significant
the split across a certain year is, the more radical the author's change in style around that year, and compare the effect of the particular year that marked the beginning
of democracy in Romania to all of the other years during which Marcus wrote his essays. We plot, in addition to the actual values of the stylistic
metrics for each year, the p-values corresponding to the statistical significance of the difference for each metric between the articles written before and after each year. 
The vertical line over the year 1990 splits the plot into two parts, corresponding to the period before respectively after the fall of communism.

\begin{figure}[htb]
\centering
\includegraphics[width=1.0\linewidth]{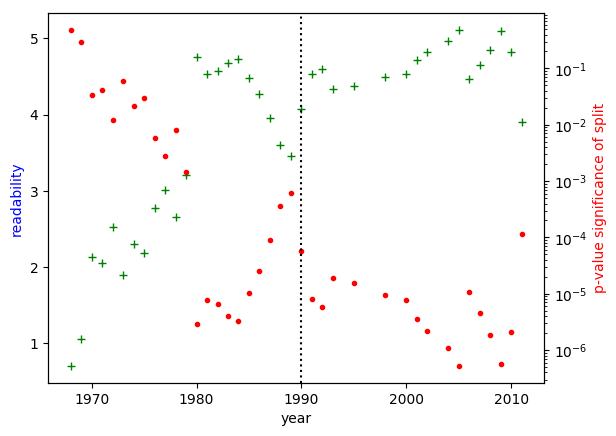}
\caption{Evolution of readability over time}
\end{figure}

\begin{figure}[htb]
\centering
\includegraphics[width=1.0\linewidth]{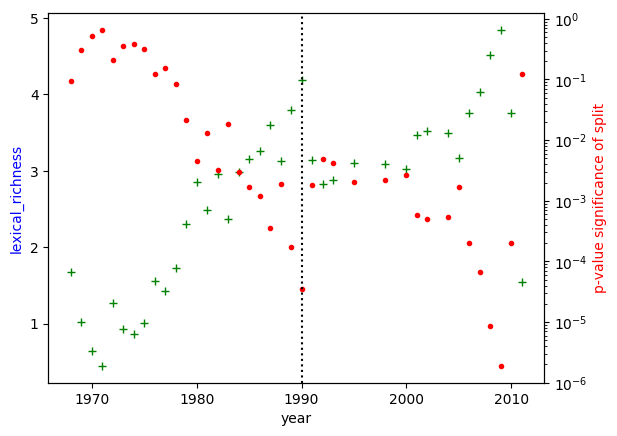}
\caption{Evolution of lexical richness over time}
\end{figure}

\begin{figure}[htb]
\centering
\includegraphics[width=1.0\linewidth]{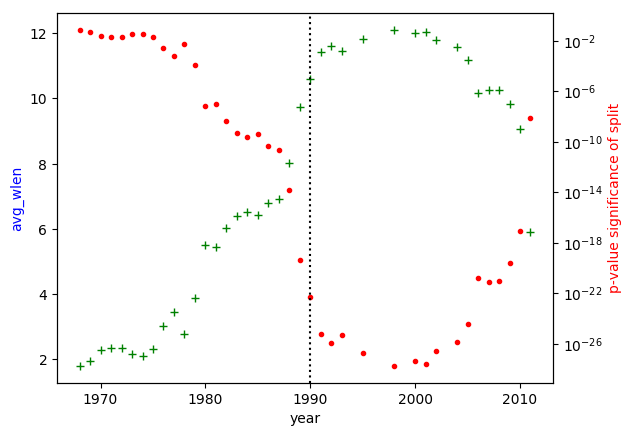}
\caption{Evolution of average word length over time}
\end{figure}

\begin{figure}[htb]
\centering
\includegraphics[width=1.0\linewidth]{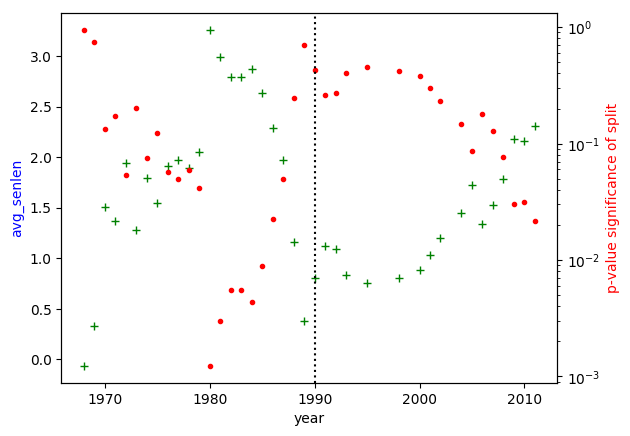}
\caption{Evolution of average sentence length over time}
\end{figure}

\section{Results and Analysis}
\label{sec-4}

The clustering experiments, as well as the 2-dimensional visualizations, show a clear distinction between the works of the two periods.
The style dendrogram in Figure 1 shows that texts tend to be grouped according to the volume they belong to, but moreover,
the volumes belonging to each period tend to be grouped together.
It is noticeable that texts belonging to the first and fourth volumes, written during the post-communist epoch (marked "RD1" and "RD4"),
are grouped together on the lower branch of the dendrogram, and the same happens with texts in volumes 2 and 3 ("RD2" and "RD3"), written
during communism, which are grouped on the upper branch.

The same tendency is visible in the 2-dimensional vector space plot of the texts. It is interesting that the separation
between the pre-communist texts (marked with green and blue points) and communist texts (marked with red and yellow points)
seems even more prominent than the separation between each of the four volumes - indicating that the main variable contributing
to style differences is indeed the period when the text was written.

A better, more detailed understanding of how exactly Marcus' style evolves can be gained from the
plots showing the variation in style from year to year. Overall, after the change of regime in 1990, we can see a slight drop in all of the stylistic features,
especially in word length and lexical richness. Most of the stylistic metrics (especially lexical richness and readability) show a more pronounced change of style right after the year 1990 - the year of the fall of the communist regime in Romania - suggesting that the change of regime did have an effect of its own on the style of the author.
An even more interesting phenomenon can be observed by looking at the variation of the p-values for splitting the article set into two arbitrary periods (before and
after each year). Except from average sentence length, for which the most drastic change seems to happen around the year 1980, for all of the other stylistic metrics
splitting the articles set into texts written under communism and under democracy is statistically significant (with p-values well under 0.05 for the year 1990,
and additionally having at least local minima around this year).

We can conclude from these results that, independently on possible variation at other levels (topic, etc),
there is inarguable variation between the texts at a stylistic level.

A similar but less obvious effect is noticeable in the content analysis. While the vector space plot of the texts again shows
a reasonably clear distinction between the texts in the two periods, the cluster analysis is less conclusive. 

\begin{table}[htb]
\centering
\begin{tabular}{ll}
Feature set & Precision\\
\hline
function words (style) & 75.80\%\\
stylistic metrics (style) & 73.12\%\\
content words (semantic content) & 74.46\%\\
topic modelling (semantic content) & 71.24\%\\
\end{tabular}\caption{\label{results1}Results of classifying texts into periods}

\end{table}

\begin{table}[htb]
\centering
\begin{tabular}{ll}
Feature set & Precision\\
\hline
function words & 61.29\%\\
content words & 84.13\%\\
\end{tabular}\caption{\label{results2}Results of classifying texts into volumes}

\end{table}

The results of the classification experiments very clearly show that essays can be easily separated into the two classes corresponding to communism and democracy,
both when using function words, stylistic metrics, and semantic features (content words, topics) as a feature set. A slightly better accuracy is obtained
for stylistic classification.
The stylistic features, in comparison to the content words, seem to be more successful
at predicting the period rather than the specific volume, indicating that style may indeed be the aspect of the texts that varies
the most between the two periods and political regimes.
Analyses of the author's change in style,
as well as statistical significance tests of the change in various stylistic metrics across the two periods, further confirm that
the change of political regime does have its own effect on the author's style.

We can conclude, given all of these results, that during the two periods, corresponding
to two different political regimes, Marcus produced works which are clearly distinct both stylistically and from the
point of view of the topics discussed. It is noteworthy that the classifier manages not only to classify the texts into
the two periods in which they were written, but is also successful at predicting the exact volume to which each essay pertains.
This suggests there may be other (volume-specific) factors contributing to the distinctiveness of the essays in each class, which
it may be interesting to further explore.

\section{Conclusions and Future Directions}
\label{sec-5}

We have proposed in this paper an analysis of the stylistic variation of the works of Solomon Marcus
between two distinct historical periods, and have presented results that show the separation
between the two periods is indeed clear at the level of the author's style.
The clustering approach based on the preference of the author regarding the functional words
shows that the functional words were unconsciously used by author in a different way in communism than in democracy period,
and we can use them to discriminate between the communist and post-communist texts of the given author.
We further confirm in several classification experiments that the texts can be automatically separated into
the two periods, and are distinguishable both at the level of style and topic.
We show the contribution of the change in political regime to the stylistic variation by looking at how the author's
style gradually changes over time, and by performing statistical significance tests on various stylistic metrics, comparing the two periods.

In future work, it would be very interesting to examine more closely how this variation is manifested.
From the stylistic point of view: what are the exact changes that occur with the passing of time and
the change of historical and cultural context? 
From the point of view of topic modelling, what are the specific changes in the topic of the text
corresponding to the two periods?
Do these changes match our intuitions about writing in the communist as compared to post-communist era?

Finally, continuing this analysis on other authors with works in both communism and post-communism, or, more generally,
across various political regimes, would be interesting to confirm the phenomenon is not unique to this author. Possibly extending this to
other authors in different cultural contexts could lead to an interesting analysis on the external factors
that influence style.

\section*{Acknowledgement}
\label{sec-6}
Liviu P. Dinu is supported by UEFISCDI, project number 53BG/2016. 

\bibliographystyle{splncs03}
\bibliography{article}{}

\begin{thebibliography}{10}
\providecommand{\url}[1]{\texttt{#1}}
\providecommand{\urlprefix}{URL }

\bibitem{baayen1996outside}
Baayen, H., Van~Halteren, H., Tweedie, F.: Outside the cave of shadows: Using
  syntactic annotation to enhance authorship attribution. Literary and
  Linguistic Computing  11(3),  121--132 (1996)

\bibitem{dinu-dinu-dumitru:2017:RANLP}
Dinu, A., Dinu, L.P., Dumitru, B.: On the stylistic evolution from communism to
  democracy: Solomon marcus study case. In: Proceedings of the International
  Conference Recent Advances in Natural Language Processing, RANLP 2017. pp.
  201--207. INCOMA Ltd., Varna, Bulgaria (September 2017),
  \url{https://doi.org/10.26615/978-954-452-049-6_028}

\bibitem{dinu2003classification}
Dinu, L.P.: On the classification and aggregation of hierarchies with ifferent
  constitutive elements. Fundamenta Informaticae  55(1),  39--50 (2003)

\bibitem{dinu2012pastiche}
Dinu, L.P., Niculae, V., {\c{S}}ulea, O.M.: Pastiche detection based on
  stopword rankings: exposing impersonators of a romanian writer. In:
  Proceedings of the Workshop on Computational Approaches to Deception
  Detection. pp. 72--77. Association for Computational Linguistics (2012)

\bibitem{dinu2008authorship}
Dinu, L.P., Popescu, M., Dinu, A.: Authorship identification of romanian texts
  with controversial paternity. In: LREC (2008)

\bibitem{kevselj2003n}
Ke{\v{s}}elj, V., Peng, F., Cercone, N., Thomas, C.: N-gram-based author
  profiles for authorship attribution. In: Proceedings of the conference
  pacific association for computational linguistics, PACLING. vol.~3, pp.
  255--264 (2003)

\bibitem{koppel2003exploiting}
Koppel, M., Schler, J.: Exploiting stylistic idiosyncrasies for authorship
  attribution. In: Proceedings of IJCAI'03 Workshop on Computational Approaches
  to Style Analysis and Synthesis. vol.~69, p.~72 (2003)

\bibitem{koppel2004authorship}
Koppel, M., Schler, J.: Authorship verification as a one-class classification
  problem. In: Proceedings of the twenty-first international conference on
  Machine learning. p.~62. ACM (2004)

\bibitem{marcus1973mathematische}
Marcus, S.: Mathematische poetik, vol. 151977. Editura Academiei (1973)

\bibitem{marcus2012rani}
Marcus, S.: Rani deschise. Editura Spandugino (2012-2017)

\bibitem{marcus1971introduzione}
Marcus, S., Nicolau, E., Stati, S.: Introduzione alla linguistica matematica.
  Bologna: Casa editrice R. P{\`a}tron (1971)

\bibitem{mendenhall1901mechanical}
Mendenhall, T.C.: A mechanical solution of a literary problem. Popular Science
  Monthly  (1901)

\bibitem{mosteller1963inference}
Mosteller, F., Wallace, D.L.: Inference in an authorship problem: A comparative
  study of discrimination methods applied to the authorship of the disputed
  federalist papers. Journal of the American Statistical Association  58(302),
  275--309 (1963)

\bibitem{popescu2008rank}
Popescu, M., Dinu, L.P.: Rank distance as a stylistic similarity. COLING
  (Posters)  8,  91--94 (2008)

\bibitem{senter1967automated}
Senter, R., Smith, E.A.: Automated readability index. Tech. rep., CINCINNATI
  UNIV OH (1967)

\bibitem{tweedie1998variable}
Tweedie, F.J., Baayen, R.H.: How variable may a constant be? measures of
  lexical richness in perspective. Computers and the Humanities  32(5),
  323--352 (1998)

\bibitem{van2005new}
Van~Halteren, H., Baayen, H., Tweedie, F., Haverkort, M., Neijt, A.: New
  machine learning methods demonstrate the existence of a human stylome.
  Journal of Quantitative Linguistics  12(1),  65--77 (2005)

\end{thebibliography}
% Emacs 24.5.1 (Org mode 8.2.10)
\end{document}